\documentclass{article}

\usepackage{arxiv}

\usepackage{comment}
\usepackage{steinmetz}
\usepackage{graphicx}
\usepackage{amsmath}
\usepackage{amssymb}
\usepackage{svg}

\usepackage{fancyhdr}
\usepackage{graphicx}
\usepackage{epstopdf}
\usepackage{mathrsfs}
\usepackage{amssymb}
\usepackage{algorithm}
\usepackage{algorithmic}
\usepackage{longtable,tabularx}
\usepackage{booktabs}					 
\usepackage{color}							
\usepackage{threeparttable}

\usepackage{verbatim}
\usepackage{bbold}
\usepackage{gensymb}
\usepackage{makecell}
\usepackage{caption}
\usepackage{subfigure}
\usepackage{wrapfig}
\usepackage{url}

\usepackage{float}
\usepackage{soul}
\graphicspath{ {./figures/} }

\title{Human-guided Swarms: Impedance Control-inspired Influence in Virtual Reality Environments}

\author{\hspace{1mm}Spencer Barclay \\
	School of Mechanical and Materials Engineering\\
	Washington State University \\
	Pullman, WA 99164, USA  \\
	\texttt{spencer.barclay@wsu.edu} \\
        \And
	\hspace{1mm}Kshitij Jerath \\
	Department of Mechanical and Industrial Engineering\\
	University of Massachusetts\\
	Lowell, MA 01854, USA \\
	\texttt{kshitij\_jerath@uml.edu} \\
}

\begin{document}
\maketitle

\begin{abstract}
Prior works in human-swarm interaction (HSI) have sought to guide swarm behavior towards established objectives, but may be unable to handle specific scenarios that require finer human supervision, variable autonomy, or application to large-scale swarms.
In this paper, we present an approach that enables human supervisors to tune the level of swarm control, and guide a large swarm using an assistive control mechanism that does not significantly restrict emergent swarm behaviors. 
We develop this approach in a virtual reality (VR) environment, using the HTC Vive and Unreal Engine 4 with AirSim plugin. 
The novel combination of an impedance control-inspired influence mechanism and a VR test bed enables and facilitates the rapid design and test iterations to examine trade-offs between swarming behavior and macroscopic-scale human influence, while circumventing flight duration limitations associated with battery-powered small unmanned aerial system (sUAS) systems.
The impedance control-inspired mechanism was tested by a human supervisor to guide a virtual swarm consisting of 16 sUAS agents. Each test involved moving the swarm's center of mass through narrow canyons, which were not feasible for a swarm to traverse autonomously. Results demonstrate that  integration of the influence mechanism enabled the successful manipulation of the macro-scale behavior of the swarm towards task completion, while maintaining the innate swarming behavior.

\end{abstract}

\section{Introduction}

As the potential for societal integration of multi-agent robotic systems increases \cite{hambling2015swarm}, the need to manage the collective behaviors of such systems also increases \cite{Jerath2015,Kim2022_CACC,kim2016mitigation}. There has been significant research effort directed towards the examination of how humans can assist in controlling such collective behaviors, such as in human-swarm interactions \cite{McMullan2019, UniversityofMelbourne2019, Peters2019}. 
Agent-agent interactions in a swarm of small unmanned aerial systems (sUAS) lead to the emergence of collective behaviors that enable effective 
coverage and exploration across large spatial extents. However, the same inherent collective behaviors can occasionally limit the ability of the sUAS swarm to focus on specific objects of interest during coverage or exploration missions \cite{Zhang2020self}. In these scenarios, the human operator or supervisor should have the opportunity to fractionally revoke or limit emergent swarm behaviors, and guide the swarm to achieve mission objectives.
For most applications, including in industry- and defense-related contexts, such human-swarm interaction (HSI) will likely require intuitive and predictable mechanisms of control to quickly translate the input of the human (such as a gesture) to an influence or effect on the sUAS swarm.

\begin{figure*}[ht]
	\centering
    \includegraphics[width=0.8\textwidth]{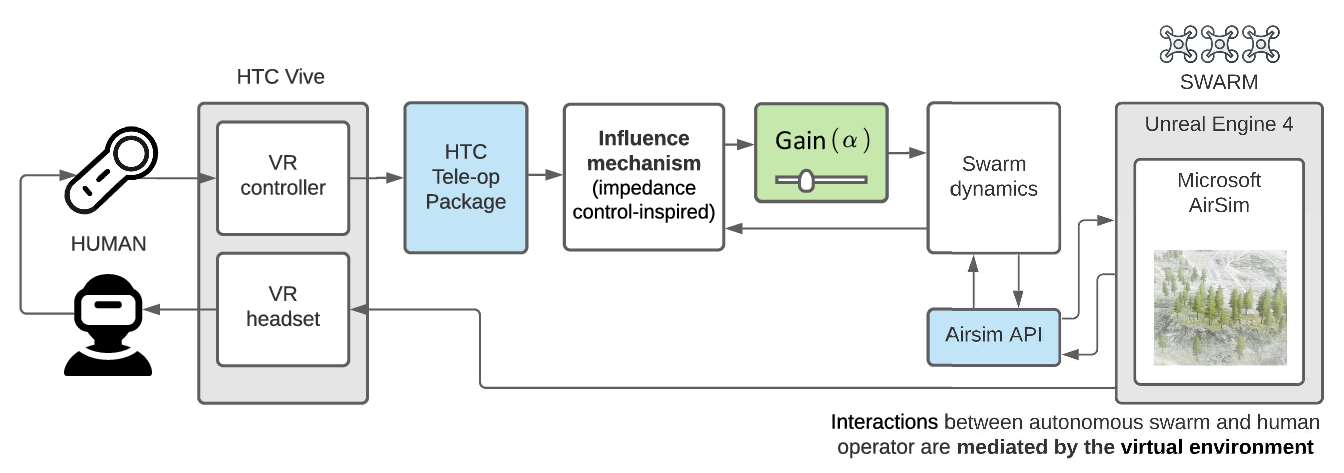}
    \caption{Feedback loop depicting flow of information across controllers and virtual reality (VR) environment. Human supervisor can provide continuous macroscopic influence to the sUAS swarm in a form of blended or shared control, if innate system behaviors fails to meet mission objectives.
    }
    \label{fig:swarm_setup}
\end{figure*}

The goal of our work is to create an intuitive interface for a human supervisor to influence or guide an sUAS swarm without excessive incursions on decentralized control afforded by these systems, while attempting to create more predictable behaviors. This is a potentially valuable approach that can enable the fully utilization of swarm capabilities, while also retaining an ongoing macroscopic-level of swarm control in scenarios where focus on specific regions of interest is required (e.g., search and rescue, surveillance operations) \cite{Scott2018}. The influence mechanism has been implemented and tested using 16 drones in a photo-realistic virtual reality (VR) environment (as shown in Fig. \ref{fig:swarm_setup}). This approach enables (a) designers to perform rapid iterations of influence mechanisms for an sUAS swarm, (b) operators to use human actions and movements as direct inputs to the swarm to potentially circumvent issues associated with short flight times of battery-powered sUAS \cite{meriaux2022evaluation, norton2022decisive}. 

The remainder of the paper is organized as follows. Section \ref{Sec:Literature_review} outlines prior relevant work on human swarm interaction, swarm dynamics, and impedance control. Section \ref{Sec:Influenceing_a_Swarm} provides a more detailed discussion of the impedance controller and human supervisor's macroscopic influence on the swarm. Section \ref{Sec:Experimental_Setup} describes the VR experimental setup and results. Finally, Section \ref{Sec:concluding_remarks} includes concluding remarks and some discussion of future works.

\section{Literature review}
\label{Sec:Literature_review}

This section discusses the state-of-the-art and its limitations with respect to impedance control and human-swarm interaction.
State-of-the-art models of emergent swarm behavior
typically define the reactive dynamics of agent interactions through either (a) continuous functions \cite{Barnes2007, Haeri2021, Tsykunov2018}, or (b) piece-wise `zone' functions \cite{CouzinNW2002, hexmoor2005swarm}. Perhaps the most widely-used swarm model is based on Couzin's work and relies on piece-wise functions that define agent interactions \cite{CouzinNW2002}.
Due to its ease of implementation and widespread utilization, our work relies on the Couzin model for generating swarm dynamics. These dynamics are discussed in more detail in Section \ref{Sec:Influenceing_a_Swarm}.

\subsection{Impedance Control}
\label{SubSec:Impedance Control}

While the reactive swarm dynamics are often discussed in a completely decentralized context, there is merit to the notion of fractional control of emergent swarming behavior. In this paper, we evaluate the use of such a human-guided fractional technique that uses impedance control. The origins of the impedance controller can be traced back to the desire to regulate the relationship between an input motion and the output force in robotic systems \cite{Hogan1989}. They were designed as a solution to dynamic control by Hogan as a result of observing the behavior of human muscles \cite{Hogan1984}\cite{Hogan1985I}. Subsequently, impedance control has been used successfully in a variety of applications. For example, hardware-implemented impedance control has been used in physically-connected systems such as robotic end effectors \cite{Bonitz1996}, and software-implemented impedance control has been used for physically-unconnected settings such as for robot obstacle avoidance \cite{Jezierski}. Impedance control is a potentially useful tool to manipulate and influence swarms, but there have been limited efforts in the past to leverage it towards this end \cite{Tsykunov2018}\cite{Engelbrecht2021adaptive}. Tsykunov et al. have demonstrated software-implemented impedance control as a means to control a swarm in response to human gestures through a wearable tactile interface \cite{Tsykunov2018}. Similarly, Engelbrecht et al. have demonstrated the use of virtual impedance control for simultaneous formation control and dynamic obstacle avoidance for a multi-agent system comprised of ground robots \cite{Engelbrecht2021adaptive}. However, these studies model the underlying agent interactions as mass-spring-damper systems. Contrarily, our work focuses on biologically-inspired agent interactions that produce collective, emergent behaviors in sUAS swarms \cite{CouzinNW2002}. We build on these swarm dynamics and leverage the advantages offered by impedance control, such as the ability to simultaneously govern the position of the swarm and the control the magnitude of force provided by the human inputs. Our approach blends human guidance with the emergent behavior of swarms via an impedance control-inspired influence mechanism, and is examined in a VR environment.

\subsection{Human-Swarm Interaction (HSI)}
\label{SubSec: HSI}
Previous research works geared towards controlling or influencing swarms have used a variety of strategies, such as utilizing a small subset of human-piloted \textit{leader} agents in conjunction with autonomous agents, or using human gestures to modify swarming behaviors \cite{Kolling2016}\cite{Nagi2014}\cite{McLurkin2006}. 
For example, Patel et al. have demonstrated the use of environment-oriented modality to allow human users to control a swarms, providing high level objectives without directly engaging robots \cite{patel2019mixed}. They also demonstrate a robot-oriented modality, where users can engage directly with individual robots. Perhaps closer in spirit to the presented work is that of Setter et al., where a haptic device is used to control team-level properties of a swarm \cite{setter2015team}. While these approaches have their advantages, teleoperation of leader agents to control a swarm may become difficult and cumbersome as the size of the swarm increases. On the other hand, current gesture-based swarm control mechanisms hinting towards global objectives may only be able to provide a coarse-level control with limited gesture vocabulary.

More recent works reflect an increasing interest in a more direct form of human guidance for swarms to enhance scalability in such applications \cite{meyerGameBenchmarkRealTime2022a}. For example, Singh et al. have designed a technique where a shepherd (human) can guide a swarm of robots to a desired objective \cite{singhModulationForceVectors2019}. While this work relies on force-based modulation to influence a swarm, its focus is largely on generating an energy-efficient mechanism for the same. In a similar approach, Macchini et al. have developed an HSI technique that enables users to directly manipulate swarms with their hands \cite{macchiniPersonalizedHumanSwarmInteraction2021}. While impressive, this work appears to limit the autonomy of the swarm in the sense that the human completely controls the motion of the swarm, eliminating the option to blend the control of the human guidance and the swarm's intrinsic behavior. A similar notion can be found in the work of Xu and Song, where they utilize reinforcement learning to enable a mixed initiative influence algorithm to learn discrete actions to be provided to the swarm \cite{xuMixedInitiativeBalance2021}. Again, while an advancement on the state-of-the-art, this works limits the ability for the human to provide more fine-grained control of the swarms in continuous space.

Overall, there is a need to generate a force-based HSI influence approach that can provide fine-grained influence of the swarm in continuous space. In this paper, we take inspiration from the advantages offered by these approaches. Specifically, we propose an influence mechanism that feeds the actions of the human supervisor through an impedance control-inspired algorithm to generate macroscopic inputs to the swarm. This continuously-applied macroscopic influence is blended with the dynamics of the autonomous swarm, to guide it in scenarios where it may be unable to achieve mission objectives on its own \cite{Desai2005blending}.

\section{Guiding Swarms with Impedance Control}
\label{Sec:Influenceing_a_Swarm}

A prerequisite to creating an influence mechanism for human-guided swarms is to recreate the underlying collective, emergent behaviors themselves \cite{Jerath2015, Haeri2021_near_optimal, jerath2014statistical, Jerath2013-pk}. In this paper, we rely on the Couzin model of swarm dynamics that relates the directional movement of each agent in the swarm to its relative positions and velocities with respect to neighboring agents \cite{CouzinNW2002}. We utilize the Couzin model in $\mathbb{R}^3$ which uses the following equations: 

\begin{align}
    \textbf{d}_r(t+\tau) &= - \sum_{j\ne i}^{n_r} \frac{\textbf{x}_{ij}(t)}{|\textbf{x}_{ij}(t)|} \label{eq:zor}\\
    \textbf{d}_o(t+\tau) &= \sum_{j=1}^{n_o} \frac{\textbf{v}_j(t)}{|\textbf{v}_j(t)|} \label{eq:zoo}\\
    \textbf{d}_a(t+\tau) &= \sum_{j\ne i}^{n_a} \frac{\textbf{x}_{ij}(t)}{|\textbf{x}_{ij}  (t)|}\label{eq:zoa}
\end{align}
where $\mathbf{x}_{ij}(t) = (\textbf{x}_j -\textbf{x}_i)/|\textbf{x}_j -\textbf{x}_i|\in \mathbb{R}^3$ represents the unit vector in the direction of neighbor $j$, $\textbf{x}_j$ represents the position vector of the $j^{th}$ agent, and $\textbf{v}_j(t)\in \mathbb{R}^3$ represents the velocity of neighboring agent $j$. Furthermore, each equation represents agent behavior corresponding to specific neighborhood zones. For example, agent $i$ will move away from neighbors that are too close, i.e. in the zone or repulsion, as described in (\ref{eq:zor}). Similarly, the response to neighboring agents in the zone of orientation is governed by their heading, as described in (\ref{eq:zoo}). Response to distant agents, i.e. agents in the zone of attraction is governed by (\ref{eq:zoa}). Each of these equations define an agent's directional vector at the next time step, given information about positions and velocities of neighboring agents. The parameters $n_r, n_o,$ and $n_a$ represent the number of agents in the zones of repulsion, orientation, and attraction at time $t$, for agent $i$, respectively. Finally, if no neighbors are detected, agent $i$ will move with the same heading as in its previous state.

If any neighbors are present in the zone of repulsion of agent $i$ (i.e. $n_r \neq 0$), repulsion is the dominant behavior. On the other hand, if no agents are present in the zone or repulsion, the orienting and attracting effects produced by more distant neighbors in the orientation zone and attraction zones will govern agent toward the agents around them. These behaviors can be represented through the following function:

\begin{equation}
    \textbf{d}_i(t+\tau) =
    \begin{cases}
                                \textbf{d}_r(t+\tau)\text{, if $n_r > 0$} \\
                                \textbf{d}_o(t+\tau)\text{, if $n_o > 0, n_r, n_a = 0$} \\
                                \textbf{d}_a(t+\tau)\text{, if $n_a > 0, n_r, n_o = 0$} \\
     \frac{1}{2}[\textbf{d}_o(t+\tau) + \textbf{d}_a(t+\tau)],\\ 
     \hspace{5em}\text{if $n_o, n_a \neq 0, n_r = 0$}\\
     \textbf{d}_i(t) \text{, otherwise}
    \end{cases}
    \label{eq:di}
\end{equation}

\subsection{Impedance Control-inspired Influence Mechanism}
With the dynamics of the autonomous sUAS swarm in place, we turn our attention towards an impedance control-inspired influence mechanism for human supervisors to guide the swarm when necessary. The influence exerted by the human supervisor manifests in a manner similar to the orientation effects modeled in the swarm dynamics. Specifically, the human supervisor holds two virtual reality (VR) controllers, one in each hand, whose positions and orientations are used to generate control inputs $\textbf{u}^{left}_i$ and $\textbf{u}^{right}_i$ (discussed later in the section). The combined effect of the actions taken by the human supervisor in moving these VR controllers  on agent $i$ is given by a single directional vector $\textbf{u}_i = \textbf{u}^{left}_i + \textbf{u}^{right}_i$. The combined effect is applied as an additive control input to every agent in the swarm by modifying the swarm dynamics shown in (\ref{eq:di}) as follows:
\begin{equation}
    \textbf{d}_i'(t+\tau) = \textbf{d}_i(t+\tau) + \alpha \textbf{u}_i(t)
\end{equation}
where $\alpha \in \mathbb{R}^+$ is a gain that represents the level of control that has been revoked from the swarm. For $\alpha=0$, the human supervisor has no influence on the original swarm dynamics, and human influence grows monotonically for successively larger values of $\alpha$. The addition of human influence to the Couzin swarming behavior is modeled as $\textbf{d}_i'(t+\tau)$ which is the new directional vector that guides agent $i$. This calculation is performed for each agent in swarm, resulting in macroscopic influence exerted by the human supervisor on the collective behavior of the swarm, while the swarm retains autonomous behaviors for low values of the gain $\alpha$.

To determine the functional form of $\textbf{u}_i(t)$, we seek inspiration from impedance control, which is a well-known and potentially viable mechanism for effecting influence. In order to manipulate a swarm via human control, such an influence mechanism should be able to modify human movements of the VR controllers into dynamical inputs for sUAS agents in the swarm. Traditionally, impedance control relates position to force, but a key realization in the current context is that the control input $\textbf{u}_i(t)$ should not represent a force, but rather a directional vector update for the agents in the swarm -- a direct consequence of the nature of the swarm dynamics. However, we can still rely on impedance control-like techniques to determine the functional form of $\textbf{u}_i(t)$, such that we can not only convert human movements into directional vector inputs for the swarming agents, but do so in a manner that can be tuned to create varying levels of fractional control.

\begin{figure}
    \centering
    \includegraphics[width=0.7\linewidth]{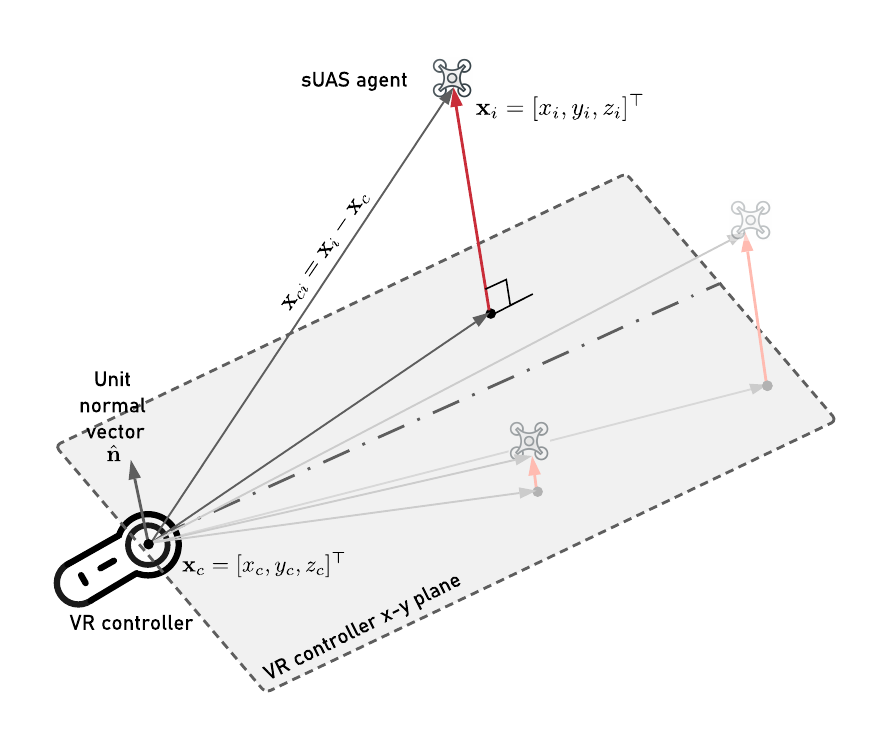}
    \caption{Impedance control-inspired influence mechanism relies on the distance of the sUAS position $\textbf{x}_i$ from the $XY$ plane (defined by the unit normal vector $\hat{\textbf{n}}$ of the VR controller) to evaluate control effort. Only one VR controller is shown in this schematic, though both are used to determine the final influence effect on the swarming agents.}
    \label{fig:Plane of operation}
\end{figure}

The goal of the impedance control-inspired influence mechanism is to increase or decrease the responsiveness  of the swarm to a change in human motion inputs. To achieve these response characteristics, we choose the following functional form, which is identical for both the left and right-handed VR controllers:
\begin{equation}
    \textbf{u}_i(t) = B_i \cdot f_B(\dot{\textbf{x}}_c,\dot{\textbf{x}}_i) + K_i \cdot f_K(\textbf{x}_c, \textbf{x}_i)
    \label{eq:Ft}
\end{equation}
where $\mathbf{x}_c \in \mathbb{R}^3$ denotes the position of the VR controller, $\textbf{x}_i \in \mathbb{R}^3$ represents the position of the agent, $K_i \in \mathbb{R}^3\times3$ denotes the `spring constants' for the stiffness between the VR controller and the sUAS agent $i$ in all three directions, and $B_i  \in \mathbb{R}^3\times3$ represents the `damping coefficients' between the VR controller and sUAS agent $i$ in all three directions. For simplicity, we assume the $K$ and $B$ matrices to be diagonal such that $K_{pp} = K$ and $B_{pp} = B$ for $p = \{1, 2, 3\}$, and $K_{pq} = B_{pq} = 0$, for $p\neq q$. Since $\mathbf{u}_i(t)$ is expected to be a unit-less directional vector rather than force, the elements of $K$ and $B$ have units of $m^{-1}$ and $(m/s)^{-1}$, respectively. Moreover, convention may suggest that the functions $f_B(\cdot, \cdot)$ and $f_K(\cdot, \cdot)$ be defined as follows: $f_B(\dot{\textbf{x}}_c,\dot{\textbf{x}}_i) = \dot{\textbf{x}}_i - \dot{\textbf{x}}_c$ and $f_K(\textbf{x}_c, \textbf{x}_i) = \textbf{x}_i - \textbf{x}_c$. However, during experiments we found that these did not yield good results and also appeared to be non-intuitive to use. Consequently, as shown in Fig. \ref{fig:Plane of operation}, the normal distance of the agent to the $XY$ plane of the VR controller was chosen to implement the controller. Thus, the control effort associated with each hand-held VR controller was evaluated as:
\begin{equation}
\textbf{u}_i(t) = B((\hat{\textbf{n}}\cdot (\dot{\textbf{x}}_i - \dot{\textbf{x}}_c))\times\hat{\textbf{n}}) + K((\hat{\textbf{n}}\cdot(\textbf{x}_i - \textbf{x}_c))\times\hat{\textbf{n}})
\label{eq:controller_final}
\end{equation}
where $\hat{\textbf{n}}$ denotes the unit normal vector to the $XY$ plane. In other words, the controller $\textbf{u}_i(t)$ compensates based on relative displacement and relative velocity between the movements of the VR controller and the agent, in the direction normal to the $XY$ plane in which the human supervisor holds the VR controllers.

The parameters in matrices $K$ and $B$ directly impact the response of the agents (and thus the swarm) to the movements enacted by the human supervisor. A stiff impedance parameter reduces the autonomous behavior of the swarm whereas softer impedance control helps retain emergent swarm behavior. Moreover, an operator may select a personalized set of parameter values for the matrices $K$ and $B$, to alter the gain $\alpha$ and modify the response towards softer (greater swarm-like behavior) or stiffer (greater human influence) response. By modifying the human input we can further tune the controller to guide the swarm towards desired objectives, such as a specific region of interest. 
In the current implementation, the gain $\alpha$ is chosen to be constant for all sUAS agents, though it is possible to differentiate agents based on varying levels of shared control as a function of the trust in agent capabilities. Of course, the value of $\alpha$ may be updated in real-time depending on the needs of the operator and the environmental scenario.
The blended or shared swarm control technique creates a quantified relationship between the two behaviors, enabling us to balance emergent swarm behaviors and macroscopic-scale human influence. Moreover, the VR implementation discussed next creates a valuable opportunity for
testing algorithms in shared control scenarios without requiring cost-prohibitive experiments.
    
\section{Experimental setup and Results}
\label{Sec:Experimental_Setup}
The experimental setup simulates a swarm of 16 drones in a virtual reality canyon environment using the Couzin model for swarm dynamics. The VR environment offers significant design and development advantages in that it enables rapid iterations during testing while incorporating the real-world physics of the game engine \cite{jerath2016simulation,scott2018mission}. Additional details are provided below. 

\subsection{Virtual Reality Setup}

\begin{figure}[!]
 \centering
 \includegraphics[width=0.6\linewidth]{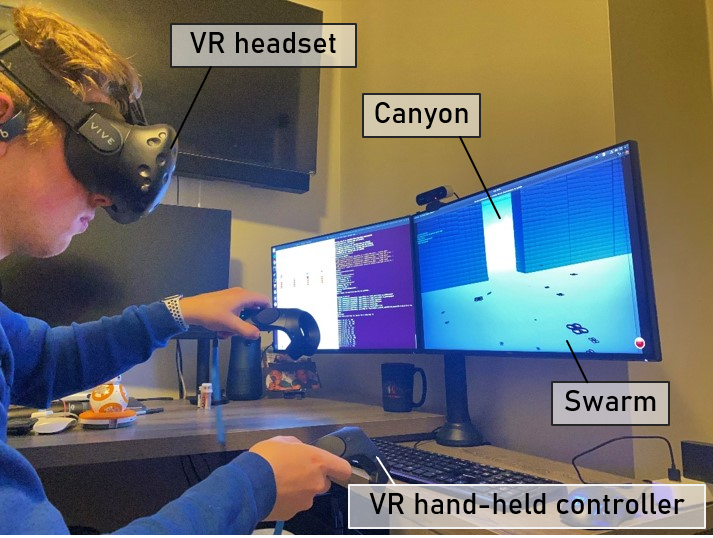}
 \caption{Human supervisor exerts macroscopic influence on autonomous swarm using hand-held VR controllers to enable it to successfully traverse a narrow canyon (visible on right monitor screen). Our virtual reality setup uses HTC Vive for control and headset to display the scene to the human supervisor, and Lighthouse for positional tracking (not pictured).}
 \label{fig:system}
 \end{figure}

The closed-loop system seen in Fig. \ref{fig:swarm_setup} shows how the different components of the implemented system interact with each other to enable the human supervisor to guide the simulated autonomous swarm. Human movements are captured using an HTC Vive system (Fig. \ref{fig:system}), which consists of (a) two Lighthouse motion-capture stations used to triangulate position and calculate orientation, (b) one headset, used to place the human into the virtual environment with the swarm, and (c) two handheld controllers, used to correlate human movements into a macroscopic influence or control input vector for the swarming agents. Each hand-held controller is used to determine a unit normal vector and a corresponding plane. Based on interactions with the two planes corresponding to the two hand-held controllers, the directional vector updates for the swarming agents are evaluated using an impedance control-inspired mechanism, as discussed in Section \ref{Sec:Influenceing_a_Swarm} and Fig. \ref{fig:Plane of operation}. 
The influence mechanism creates a unique output for every agent in the swarm. The Vive hand-held VR controllers provide accurate and precise output of human movements, enabling effective and intuitive manipulation of the swarm.

The study also uses the Unreal 4 gaming engine, along with the Microsoft Airsim plugin, to generate realistic physics and ensure accurate test results. The Unreal Engine 4 and AirSim environment enables continuous swarm testing of robotic vehicles and provides detailed sensor information from each sUAS, while circumventing flight duration limitations associated with battery powered sUAS systems \cite{airsim2017fsr}.

 \begin{figure}[!]
\centering
    \includegraphics[width=0.7\linewidth]{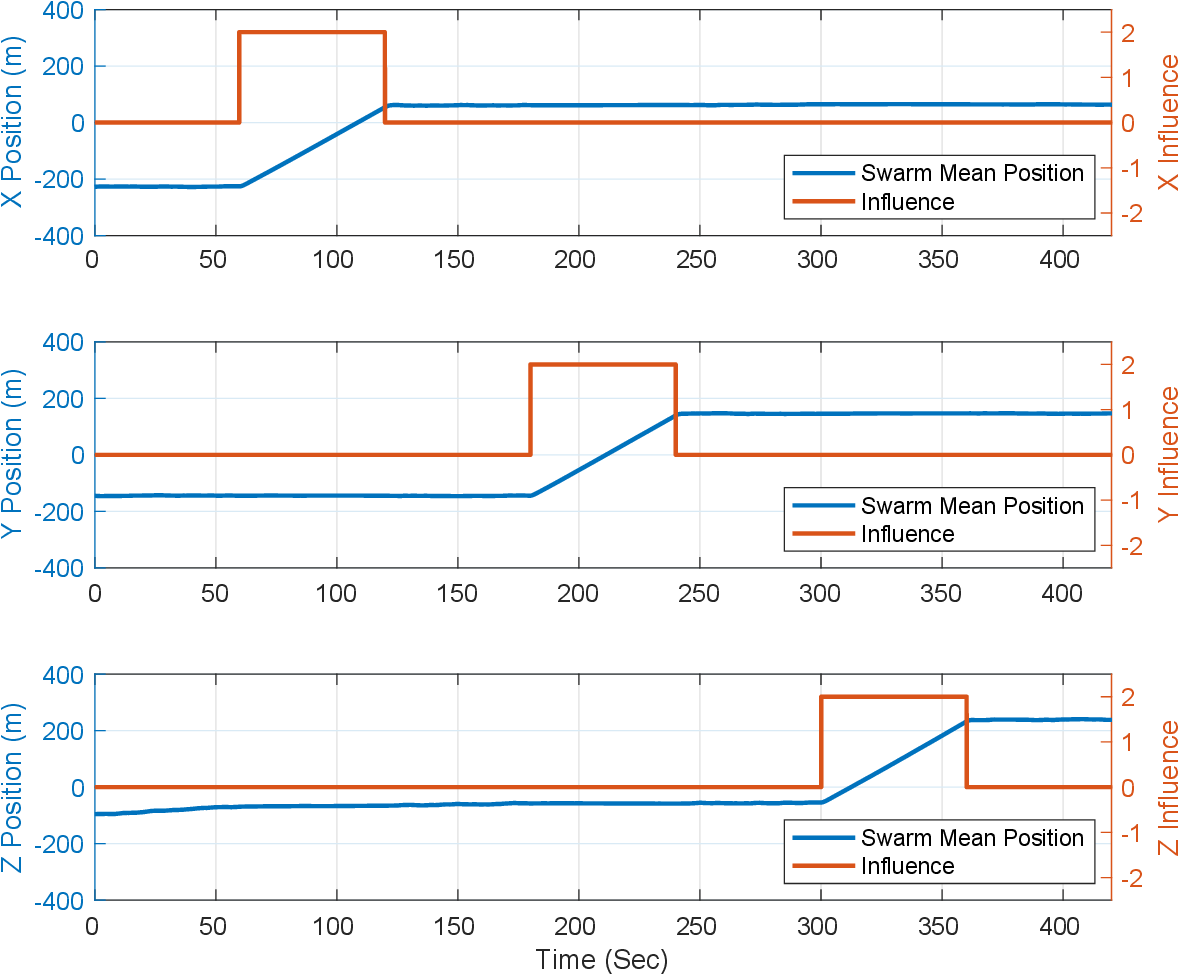}
    \caption{Algorithmic application of pulse input-like influence along a single axis and its effect on the the swarm's movement along that axis. Programatically-applied inputs demonstrate the intended effects of the influence mechanism in a scenario \textit{without explicit} human influence $(\alpha = 5)$.
    }
    \label{fig:SwarmProgInfluence}
\end{figure}

\subsection{Results}
\label{SubSec:Re}
To examine if the devised influence mechanism generates desired results, we first test the swarm behavior with the human supervisor inputs replace with a pre-determined algorithmically applied input, with gain of $\alpha=5$, and with damping and stiffness matrices, $B = 0.5\textbf{I}$ and $K = \textbf{I}$, respectively, where $\textbf{I}$ represents the identity matrix. Alternative values of $B$ and $K$ can be chosen to personalize the influence mechanism to match the human supervisor. As evident from Fig. \ref{fig:SwarmProgInfluence}, the algorithmic application of a pulse-like influence at various time instances generates the intended macroscopic effects in terms of impacting the mean position of the swarm. Specifically, applying influence in one direction causes the swarm to move in the same direction at an aggregate level.

\begin{figure}
    \centering
    \includegraphics[width=0.6\linewidth]{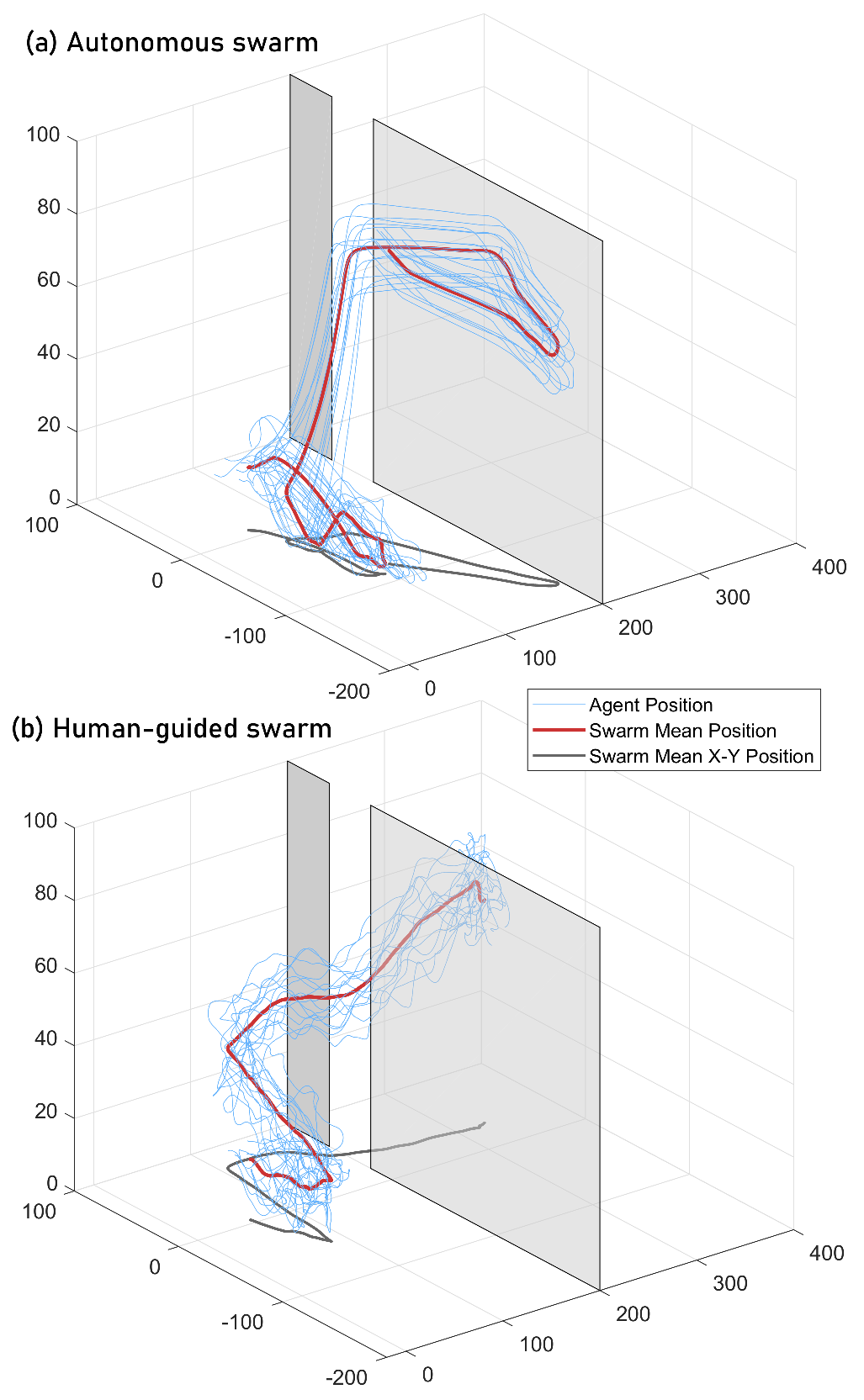}
    \caption{Agent positions and swarm mean position: (a) with autonomous operation in cohesive flight $(\alpha = 0)$, and (b) guided by human supervisor during milling (with gain $\alpha=5$). Grey vertical planes represent walls, with the narrow gap representing a canyon. Dark gray line denotes projection of mean swarm position onto the $XY$ ground plane.}
    \label{fig:3DSwarm}
\end{figure}

Next, we compare the performance of the autonomous swarm in passing through a canyon environment, which is represented in the virtual environment as a narrow gap in a wall (Fig. \ref{fig:system}). As seen in Fig. \ref{fig:3DSwarm}(a), the autonomous swarm in parallel or cohesive flight (with gain $\alpha = 0$) is unable to traverse the canyon. The dark grey line in the figure represents the projection of the mean swarm position onto the $XY$ ground plane, which shows that the swarm remains in the foreground of the wall. 
On the other hand, as seen in Fig. \ref{fig:3DSwarm}(b), a human supervisor is able to use the influence mechanism enabled by the VR controller to manipulate the swarm during milling behavior, alter its aggregate orientation, and fly it through the narrow gap. It is notable that in this operation, the swarm is allowed to retain some level of autonomous behavior as  governed by the underlying swarm dynamics. Thus, the human supervisor is able to effect macroscopic influence on the swarm without significantly altering the agent-agent interactions occurring at the microscopic level.

\begin{figure}[h]
\centering
    \includegraphics[width=0.6\linewidth]{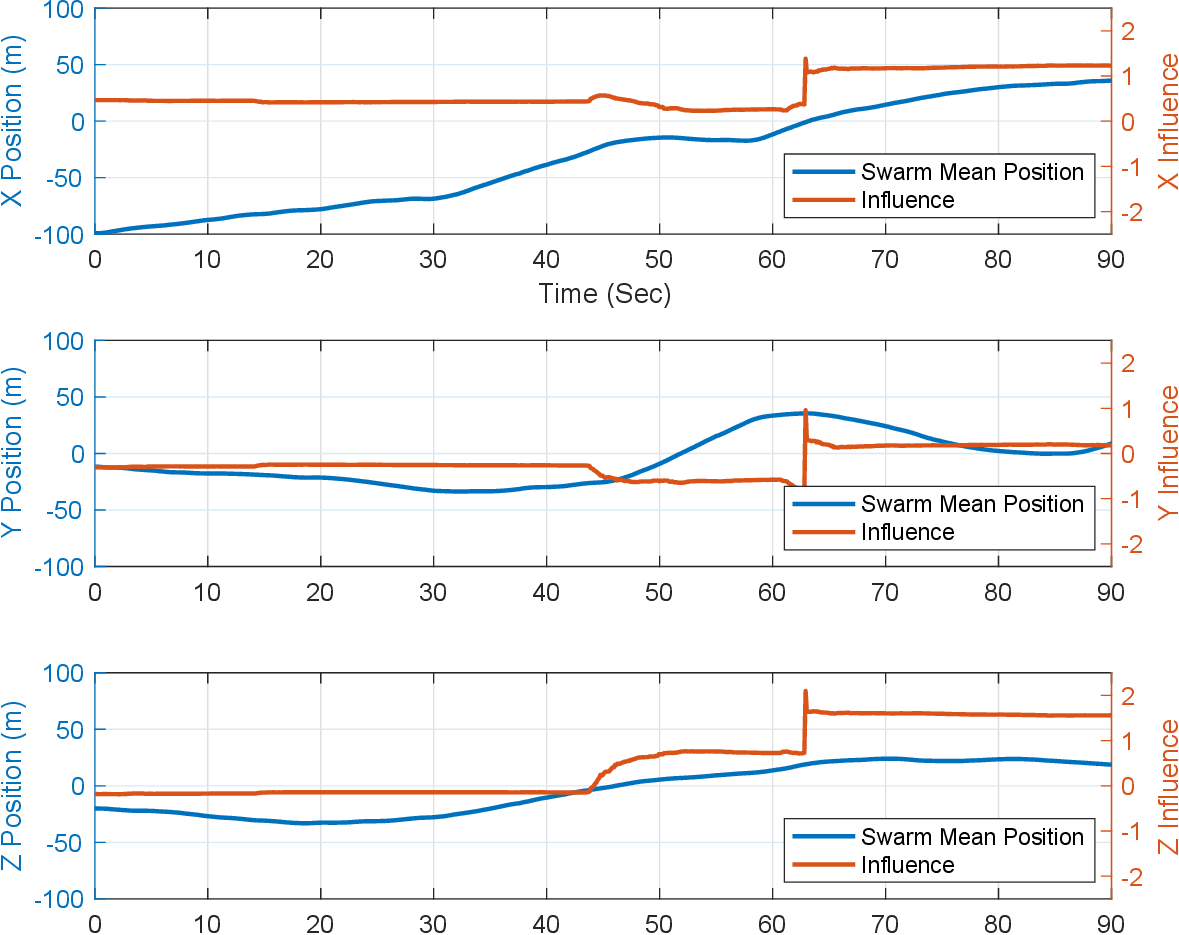}
    \caption{The  comparison  between  influence  on  a  single  axis  and  its effect on the the swarm's movement along that axis. 
    }
    \label{fig:SwarmInfluence}
\end{figure}

\begin{figure}[h]
\centering
    \includegraphics[width=0.6\linewidth]{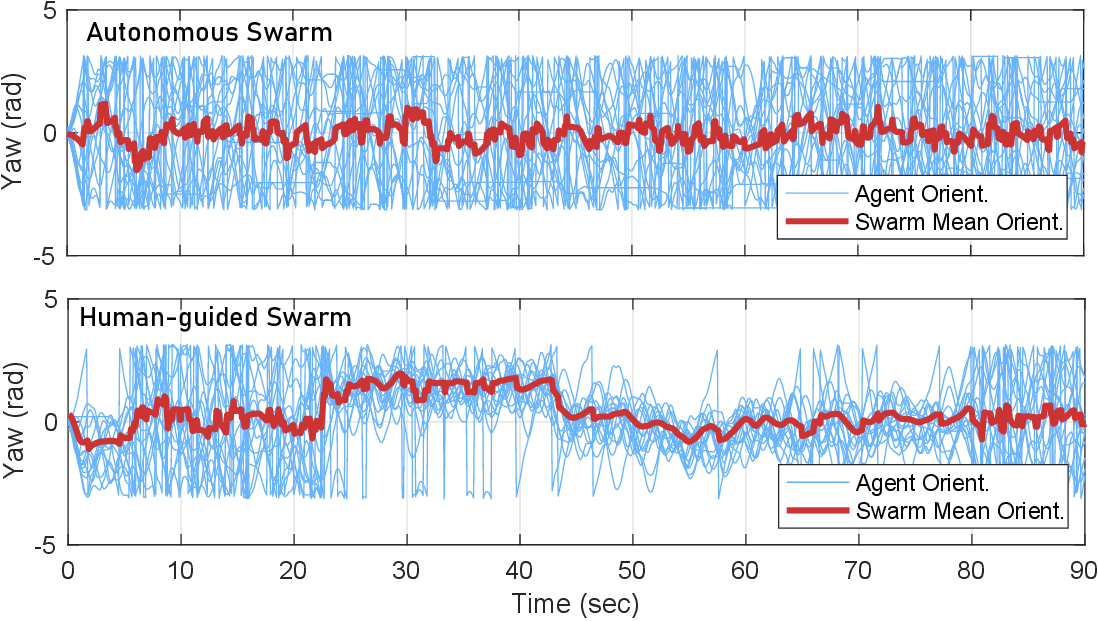}
    \caption{The average yaw of the swarm exhibiting milling behavior while (a) operating autonomously $(\alpha=0)$, and (b) being influenced by the human supervisor $(\alpha=5)$. Dynamics in roll and pitch are nearly identical for both scenarios and not shown.
    }
    \label{fig:NOVRSwarmOrient}
\end{figure}

Fig. \ref{fig:SwarmInfluence} shows the subtle influence operations executed by the human supervisor to alter the mean positions of the swarm. In addition, observing the orientation of the agents and the aggregate swarm also demonstrate the ability of the influence mechanism to have the intended effects on the swarm. For example, when the swarm is exhibiting milling behavior, the average yaw of the swarm hovers around zero. However, when the human supervisor engages in some for of influence, the average yaw can clearly be seen to depart from the zero mean and towards a specific heading, as shown in Fig. \ref{fig:NOVRSwarmOrient}. The dynamics of the swarming agents in roll and pitch remain largely unchanged, indicating that some measure of swarm dynamics are retained even when the human supervisor is exerting macroscopic influence on the system.    


\section{Concluding remarks and future works}
\label{Sec:concluding_remarks}
This paper presented a novel combination of an impedance control-inspired influence mechanism with a VR experimental setup to enable rapid design and testing of control policies that enable human supervisors to exert macroscopic influence on an sUAS swarm.  The presented approach to influencing a swarm utilizes the useful characteristics of both swarm dynamics and human control, while incorporating elements of blended or shared control. Additionally, the use of the impedance control-inspired influence mechanism helps mitigate potential adverse effect on swarm operation due sudden movements of human supervisor or operator fatigue. 
As demonstrated via the canyon problem (Fig. \ref{fig:3DSwarm}), the macroscopic influence exerted by the human supervisor provides the swarm the ability to navigate through difficult and potentially impassable obstacles. 
Adding the notion of impedance control and fractional gain may provide the human supervisor the ability to adapt the swarm to different environments and change the effect influence on the swarm, with limited effects on the underlying agent-agent interactions that drive collective, emergent behaviors.

The VR experimental setup also provides the opportunity to explore future avenues for research.
A natural next step is to evaluate the performance of the human-guided swarms with multiple human participants. This will also provide the opportunity to evaluate the distribution of parameter values in $B$ and $K $matrices as a means to personalize the influence mechanism for different individuals.
Another logical step for continuing this research is implementing the controller with a real-world swarm with similar swarming characteristics, as well as potentially using an augmented reality system to control the physical swarm.

\bibliographystyle{ieeetr}
\bibliography{human_guided_swarms}

\end{document}